\pdfoutput=1
\documentclass[11pt]{article}
\usepackage[table]{xcolor} 

\usepackage{eacl2023}
\usepackage{multirow}

\usepackage{times}
\usepackage{latexsym}
\usepackage{verbatim}

\usepackage{enumitem}
\usepackage[T1]{fontenc}
\usepackage[utf8]{inputenc}

\usepackage{microtype}

\usepackage{adjustbox}
\usepackage{tikz}
\usetikzlibrary{shapes,positioning}

\usepackage{hyperref}

\usepackage{array}
\newcolumntype{C}[1]{>{\centering\arraybackslash}p{#1}}

\definecolor{propertyshade}{HTML}{EFEFEF}

\title{Missing Information, Unresponsive Authors, Experimental Flaws:\\The Impossibility of Assessing the Reproducibility\\of Previous Human Evaluations in NLP\vspace{0.2cm}}

\author{Anya Belz$^{a,b}$\normalfont{\texttt{(anya.belz@adaptcentre.ie)}}\textbf{\hspace{-0.075cm}, Craig Thomson}$^b$\textbf{,} \textbf{Ehud Reiter}$^b$\textbf{,}\\
  \textbf{Gavin Abercrombie$^8$, Jose M.\ Alonso-Moral$^{17}$, Mohammad Arvan$^{16}$, Anouck Braggaar$^{13}$,}\\\textbf{Mark Cieliebak$^{20}$, Elizabeth Clark$^6$, Kees van Deemter$^{19}$, Tanvi Dinkar$^8$, Ondřej Dušek$^9$,}\\ \textbf{Steffen Eger$^1$, Qixiang Fang$^{19}$, Mingqi Gao$^{11}$, Albert Gatt$^{19}$, Dimitra Gkatzia$^4$, Javier}\\\textbf{González-Corbelle$^{17}$, Dirk Hovy$^2$,  Manuela Hürlimann$^{20}$, Takumi Ito$^{10}$, John D.\ Kelleher$^{12}$,}\\\textbf{ Filip Klubička$^{12}$, Emiel Krahmer$^{13}$, Huiyuan Lai$^7$, Chris van der Lee$^{13}$, Yiru Li$^7$, Saad}\\\textbf{Mahamood$^{14}$, Margot Mieskes$^{15}$, Emiel van Miltenburg$^{13}$, Pablo Mosteiro$^{19}$, Malvina}\\\textbf{Nissim$^7$, Natalie Parde$^{16}$, Ondřej Plátek$^9$, Verena Rieser$^8$, Jie Ruan$^{11}$, Joel Tetreault$^3$,}\\\textbf{Antonio Toral$^7$, Xiaojun Wan$^{11}$, Leo Wanner$^{18}$, Lewis Watson$^4$, Diyi Yang$^5$} \vspace{.24cm} \\
  $^a$ADAPT/DCU, Ireland; $^b$University of Aberdeen, UK;  $^1$Bielefeld University, Germany; $^2$Bocconi\\ University, Italy; $^3$Dataminr, US; $^4$Edinburgh Napier University, UK; $^5$Georgia Tech, US; $^6$Google \\Research, US; $^7$Groningen University, Netherlands; $^8$Heriot-Watt University, UK; $^9$Charles University\\ Prague, Czechia; $^{10}$Tohoku University, Japan; $^{11}$Peking University, China; $^{12}$Technological\\ University Dublin, Ireland; $^{13}$Tilburg University, Netherlands; $^{14}$trivago, Germany; $^{15}$University of\\ Applied Sciences Darmstadt, Germany; $^{16}$University of Illinois Chicago, US; $^{17}$Universidade de\\ Santiago de Compostela, Spain; $^{18}$Universitat Pompeu Fabra, Spain; $^{19}$Utrecht University,\\Netherlands; $^{20}$Zurich University of Applied Sciences, Switzerland
}

\begin{document}

\maketitle

\begin{abstract}
We report our efforts in identifying a set of previous human evaluations in NLP that would be suitable for a coordinated study examining what makes human evaluations in NLP more/less reproducible.
We present our results and findings, which include that  just 13\% of papers had (i) sufficiently low barriers to reproduction, and (ii) enough obtainable information, to be considered for reproduction, 
and that all but one of the experiments we selected for reproduction was discovered to have flaws that made the meaningfulness of conducting a reproduction questionable. As a result, we had to change our coordinated study design from a reproduce approach to a standardise-then-reproduce-twice approach. Our overall (negative) finding that the great majority of human evaluations in NLP is not repeatable and/or not reproducible and/or too flawed to justify reproduction, paints a dire picture, but presents an opportunity for a rethink about how to design and report human evaluations in NLP. 
\end{abstract}

\section{Introduction}\label{sec:intro}

There is increasing awareness in Natural Language Processing (NLP) that reproducibility of results, most particularly of results from system evaluations, matters greatly, and that currently the field does not assess reproducibility of results rigorously enough, and lacks a common approach to it. Recent work has made progress particularly with respect to automatic evaluation \cite{pineau2020checklist, whitaker2017}, but reproducibility of human evaluation, widely considered the litmus test of quality in NLP, has received less attention. It could be argued that if it is not known how reproducible human evaluations are, it is not known how reliable they are; and if it is not known how reliable they are, then it is not known how reliable automatic evaluations meta-evaluated against them 
are either. 

The work reported in this paper forms part of the ReproHum project\footnote{\url{https://gow.epsrc.ukri.org/NGBOViewGrant.aspx?GrantRef=EP/V05645X/1}} in which our aim is to build on existing work on recording properties of human evaluations datasheet-style \cite{shimorina-belz-2022-human}, and assessing how close results from a reproduction study are to the original study \cite{belz-etal-2022-quantified}, to investigate systematically what factors make a human evaluation more---or less---reproducible. In this paper, we present the findings from our work on the project so far which necessitated a rethink of our entire approach to designing such an investigation.

Section~\ref{sec:motivation} outlines our motivation for carrying out a multi-lab multi-test (MLMT) study of factors affecting reproduciblity in NLP, and our original design for the study. Section~\ref{sec:sel-anno} describes our paper selection, annotation and filtering process which yielded a surprisingly small number of candidate papers for reproduction. In Section~\ref{sec:flaws} we describe the numerous further issues with original evaluation studies we encountered in the process of setting up reproductions of them with partner labs. Section~\ref{sec:conclusion} summarises our negative findings regarding the infeasibilty of assessing the reproducibility of previously conducted human evaluations in NLP as they are, and outlines the changes to our multi-lab multi-test study necessitated by the findings.

\section{Motivation and Overall Study Design}\label{sec:motivation}

Individual studies can tell us how close a reproduction study's results are to those in the original study. A large number of such studies can show  general tendencies regarding what kinds of evaluations have better reproducibility.
However, we do not currently have a large number of reproduction studies in NLP and because of their cost and  lack of appeal, this is unlikely to change. Moreover, accumulations of individual studies do not provide the conditions in which the effect size and significance of specific factors on reproducibility, and interactions between them, can be measured. 

To create such conditions, a controlled study of equal numbers of reproductions with and without factors of interest is needed. Moreover, we know from existing work \cite{belz-etal-2022-quantified,huidrom-etal-2022-two} that different reproductions of the \textit{same} original work can produce very different results. Finally, while it is instructive to test for reproducibility under identical conditions, it is also of interest to test how far good reproducibility can stretch -- e.g.\ is reproducibility affected by replacing, say, a 7-point quality scale with a 5-point one. 

A study of factors that increase/decrease reproducibility  therefore needs to (i) conduct more than one reproduction of each original study, (ii) carried out by a good mix of different teams, and to (iii) incorporate multiple rounds with decreasing similarity of conditions. The steps in setting up such a study would be as follows:

\vspace{-.1cm}

\begin{enumerate}[itemsep=0pt]
    \item Identifying candidate evaluation experiments from which to select experiments with balanced factors to include in the MLMT study;
    \item Recording properties of evaluation experiments to make it possible to select factors and control for them;
    \item Selecting factors to control for and corresponding subsets of experiments; and
    \item Carrying out reproductions for the selected evaluation studies and factors. % 
\end{enumerate}

\noindent We describe Steps~1 and~2 in Sections~\ref{ssec:filtering} and~\ref{ssec:props}, Step~3 in \ref{ssec:selecting-props}, and Step~4 up to the point where we  aborted the original study design in Section~\ref{sec:flaws}.

\section{Selection and Assessment of Candidate Evaluation Experiments}\label{sec:sel-anno}

\begin{figure}
    \begin{adjustbox}{max width=\columnwidth}
        % Graphic for TeX using PGF
        % Title: /home/badger/eacl.dia
        % Creator: Dia v0.97+git
        % CreationDate: Thu Dec  8 09:57:58 2022
        % \usepackage{tikz}
        % The following commands are not supported in PSTricks at present
        % We define them conditionally, so when they are implemented,
        % this pgf file will use them.
        \ifx\du\undefined
          \newlength{\du}
        \fi
        \setlength{\du}{15\unitlength}

        \begin{tikzpicture}[even odd rule]
            \pgftransformxscale{1.000000}
            \pgftransformyscale{-1.000000}
            \definecolor{dialinecolor}{rgb}{0.000000, 0.000000, 0.000000}
            \definecolor{sidefillcolor}{HTML}{90bbfb}
        	\definecolor{diafillcolor}{HTML}{FFFFFF}
            \pgfsetfillopacity{1.000000}
            \pgfsetlinewidth{0.100000\du}
            \pgfsetstrokecolor{dialinecolor}
            \pgfsetstrokeopacity{1.000000}
            \pgfsetfillcolor{diafillcolor}
            \pgfsetdash{}{0pt}
            \pgfsetarrowsend{stealth}

            \pgfsetmiterjoin
            {\pgfsetcornersarced{\pgfpoint{0.000000\du}{0.000000\du}}}

            % One Box
            \draw (15.000000\du,5.000000\du)--(15.000000\du,10.000000\du)--(25.000000\du,10.000000\du)--(25.000000\du,5.000000\du)--cycle;
            \node[anchor=base,inner sep=0pt, outer sep=0pt,color=dialinecolor] at (20.000000\du,6.495000\du){ACL \& TACL Papers (p)};
            \node[anchor=base,inner sep=0pt, outer sep=0pt,color=dialinecolor] at (20.000000\du,7.295000\du){identified through};
            \node[anchor=base,inner sep=0pt, outer sep=0pt,color=dialinecolor] at (20.000000\du,8.095000\du){ACL Anthology search};
            \node[anchor=base,inner sep=0pt, outer sep=0pt,color=dialinecolor] at (20.000000\du,8.895000\du){(p=177)};

            % One Box
            \draw (15.000000\du,11.600000\du)--(15.000000\du,15.100000\du)--(25.000000\du,15.100000\du)--(25.000000\du,11.600000\du)--cycle;
            \node[anchor=base,inner sep=0pt, outer sep=0pt,color=dialinecolor] at (20.000000\du,12.745000\du){Papers after manually};
            \node[anchor=base,inner sep=0pt, outer sep=0pt,color=dialinecolor] at (20.000000\du,13.545000\du){checked for suitability.};
            \node[anchor=base,inner sep=0pt, outer sep=0pt,color=dialinecolor] at (20.000000\du,14.345000\du){(p=116)};

            % One Box
            \draw (28.000000\du,12.500000\du)--(28.000000\du,14.400000\du)--(37.950000\du,14.400000\du)--(37.950000\du,12.500000\du)--cycle;
            \node[anchor=base,inner sep=0pt, outer sep=0pt,color=dialinecolor] at (32.975000\du,13.645000\du){Papers excluded (p=61)};

            % One Box
            \draw (15.000000\du,16.500000\du)--(15.000000\du,21.500000\du)--(25.000000\du,21.500000\du)--(25.000000\du,16.500000\du)--cycle;
            \node[anchor=base,inner sep=0pt, outer sep=0pt,color=dialinecolor] at (20.000000\du,18.395000\du){Papers where};
            \node[anchor=base,inner sep=0pt, outer sep=0pt,color=dialinecolor] at (20.000000\du,19.195000\du){corresponding author};
            \node[anchor=base,inner sep=0pt, outer sep=0pt,color=dialinecolor] at (20.000000\du,19.995000\du){responded (p=45)};

            % One Box
            \fill (28.000000\du,18.000000\du)--(28.000000\du,20.000000\du)--(38.000000\du,20.000000\du)--(38.000000\du,18.000000\du)--cycle;
            \draw (28.000000\du,18.000000\du)--(28.000000\du,20.000000\du)--(38.000000\du,20.000000\du)--(38.000000\du,18.000000\du)--cycle;
            \node[anchor=base,inner sep=0pt, outer sep=0pt,color=dialinecolor] at (33.000000\du,19.195000\du){No response (p=71)};

            % One Box
            \draw (15.000000\du,23.000000\du)--(15.000000\du,27.000000\du)--(25.000000\du,27.000000\du)--(25.000000\du,23.000000\du)--cycle;
            \node[anchor=base,inner sep=0pt, outer sep=0pt,color=dialinecolor] at (20.000000\du,24.395000\du){Papers where author};
            \node[anchor=base,inner sep=0pt, outer sep=0pt,color=dialinecolor] at (20.000000\du,25.195000\du){indicated that details};
            \node[anchor=base,inner sep=0pt, outer sep=0pt,color=dialinecolor] at (20.000000\du,25.995000\du){could be provided (p=20)};

            % One Box
            \draw (28.000000\du,21.500000\du)--(28.000000\du,24.200000\du)--(38.000000\du,24.200000\du)--(38.000000\du,21.500000\du)--cycle;
            \node[anchor=base,inner sep=0pt, outer sep=0pt,color=dialinecolor] at (33.000000\du,22.645000\du){Author could not send};
            \node[anchor=base,inner sep=0pt, outer sep=0pt,color=dialinecolor] at (33.000000\du,23.445000\du){outputs/interface (p=16)};

            % One Box
            \draw (28.000000\du,25.500000\du)--(28.000000\du,28.200000\du)--(38.000000\du,28.200000\du)--(38.000000\du,25.500000\du)--cycle;
            \node[anchor=base,inner sep=0pt, outer sep=0pt,color=dialinecolor] at (33.000000\du,26.645000\du){Correspondence stalled}; 
            \node[anchor=base,inner sep=0pt, outer sep=0pt,color=dialinecolor] at (33.000000\du,27.445000\du){(p=8)};

            % One Box
            \draw (28.000000\du,29.500000\du)--(28.000000\du,32.200000\du)--(38.000000\du,32.200000\du)--(38.000000\du,29.500000\du)--cycle;
            \node[anchor=base,inner sep=0pt, outer sep=0pt,color=dialinecolor] at (33.000000\du,31.045000\du){Papers excluded (p=1)};            
            
            % One Box
            \draw (15.000000\du,28.500000\du)--(15.000000\du,32.500000\du)--(25.000000\du,32.500000\du)--(25.000000\du,28.500000\du)--cycle;
            \node[anchor=base,inner sep=0pt, outer sep=0pt,color=dialinecolor] at (20.000000\du,29.895000\du){Papers split into};
            \node[anchor=base,inner sep=0pt, outer sep=0pt,color=dialinecolor] at (20.000000\du,30.695000\du){individual experiments (e)};
            \node[anchor=base,inner sep=0pt, outer sep=0pt,color=dialinecolor] at (20.000000\du,31.495000\du){(e=28, from p=20)};

            % One Box
            \draw (15.000000\du,34.000000\du)--(15.000000\du,38.000000\du)--(25.000000\du,38.000000\du)--(25.000000\du,34.000000\du)--cycle;
            \node[anchor=base,inner sep=0pt, outer sep=0pt,color=dialinecolor] at (20.000000\du,35.395000\du){Experiments successfully};
            \node[anchor=base,inner sep=0pt, outer sep=0pt,color=dialinecolor] at (20.000000\du,36.195000\du){annotated for all factors};
            \node[anchor=base,inner sep=0pt, outer sep=0pt,color=dialinecolor] at (20.000000\du,36.995000\du){(e=20, from p=15)};

            % One Box
            \draw (28.000000\du,34.500000\du)--(28.000000\du,37.500000\du)--(38.045000\du,37.500000\du)--(38.045000\du,34.500000\du)--cycle;
            \node[anchor=base,inner sep=0pt, outer sep=0pt,color=dialinecolor] at (33.022500\du,35.795000\du){Experiments with missing};
            \node[anchor=base,inner sep=0pt, outer sep=0pt,color=dialinecolor] at (33.022500\du,36.595000\du){details (e=8, from p=5)};

            % Arrows
            \pgfsetbuttcap
            {
            	\draw (20.000000\du,10.048486\du)--(20.000000\du,11.600000\du);
            }
            
            \pgfsetbuttcap
            {
            	\draw (20.000000\du,15.100000\du)--(20.000000\du,16.500000\du);
            }
            
            \pgfsetbuttcap
            {
            	\draw (20.000000\du,21.500000\du)--(20.000000\du,22.962891\du);
            }
            
            \pgfsetbuttcap
            {
            	\draw (25.050145\du,13.388922\du)--(27.950197\du,13.411273\du);
            }
            
            \pgfsetbuttcap
            {
            	\draw (25.049957\du,19.000000\du)--(27.950043\du,19.000000\du);
            }

            \pgfsetbuttcap
            {
            	\draw (25.049957\du,25.000000\du)--(26.500000\du,25.000000\du)--(26.500000\du,22.850000\du)--(27.950043\du,22.850000\du);
            }
            
            \pgfsetbuttcap
            {
            	\draw (25.000000\du,25.000000\du)--(26.500000\du,25.000000\du)--(26.500000\du,26.850000\du)--(27.950043\du,26.850000\du);
            }

            \pgfsetbuttcap
            {
            	\draw (25.000000\du,25.000000\du)--(26.500000\du,25.000000\du)--(26.500000\du,30.800000\du)--(27.950043\du,30.800000\du);
            }            
        
            \pgfsetbuttcap
            {
            	\draw (20.000000\du,27.000000\du)--(20.000000\du,28.500000\du);
            }
            
            \pgfsetbuttcap
            {
            	\draw (20.000000\du,32.500000\du)--(20.000000\du,34.000000\du);
            }

            \pgfsetbuttcap
            {
                \draw (25.000000\du,36.000000\du)--(28.000000\du,36.000000\du);
            }

            % Side bar
            \pgfsetfillcolor{sidefillcolor}
            {
                \pgfsetcornersarced{\pgfpoint{0.400000\du}{0.400000\du}}
                \fill (10.500000\du,5.000000\du)--(10.500000\du,14.000000\du)--(12.500000\du,14.000000\du)--(12.500000\du,5.000000\du)--cycle;
                \draw (10.500000\du,5.000000\du)--(10.500000\du,14.000000\du)--(12.500000\du,14.000000\du)--(12.500000\du,5.000000\du)--cycle;
            }
            \node[anchor=base,inner sep=0pt, outer sep=0pt, rotate=90, color=dialinecolor] at (11.650000\du,9.500000\du){Search \& paper annotation};

            {
                \pgfsetcornersarced{\pgfpoint{0.400000\du}{0.400000\du}}
                \fill (10.500000\du,18.000000\du)--(10.500000\du,26.000000\du)--(12.500000\du,26.000000\du)--(12.500000\du,18.000000\du)--cycle;
                \draw (10.500000\du,18.000000\du)--(10.500000\du,26.000000\du)--(12.500000\du,26.000000\du)--(12.500000\du,18.000000\du)--cycle;
            }
            \node[anchor=base,inner sep=0pt, rotate=90, outer sep=0pt,color=dialinecolor] at (11.650000\du,22.000000\du){Author correspondence};
            
            {
                \pgfsetcornersarced{\pgfpoint{0.400000\du}{0.400000\du}}
                \fill (10.500000\du,29.000000\du)--(10.500000\du,38.000000\du)--(12.500000\du,38.000000\du)--(12.500000\du,29.000000\du)--cycle;
                \draw (10.500000\du,29.000000\du)--(10.500000\du,38.000000\du)--(12.500000\du,38.000000\du)--(12.500000\du,29.000000\du)--cycle;
            }
            \node[anchor=base,inner sep=0pt, rotate=90, outer sep=0pt,color=dialinecolor] at (11.650000\du,33.500000\du){Experiment-level annotation};
            
        \end{tikzpicture}
    \end{adjustbox}
    \caption{Flow diagram of the paper selection process, showing the decreasing number of papers that were suitable as more information was sought.}
    \label{fig:flow_diagram}
\end{figure}

Figure~\ref{fig:flow_diagram} shows the selection and annotation process in the form of a flow diagram showing the decreasing number of remaining papers/experiments. The first step was to conduct a search on the ACL Anthology for papers published in ACL (main conference) or TACL in the 2018--2022 period\footnote{Search performed in July 2022, so some TACL papers from later that year are not included.} which included the phrases ``human evaluation'' and ``participants;'' we found 177 such papers.

\subsection{High-level paper annotation}\label{ssec:filtering}

In a first round of annotating papers with properties of human evaluations, we used the following paper-level properties, annotated using only information from the paper or supplementary material: % 

\begin{enumerate}[itemsep=0pt]
    \item How many systems were evaluated;
    \item How many datasets were used;
    \item Type of participant (e.g.\ MTurk);
    \item How many unique participants;
    \item Rough estimate of how many judgments;
    \item Type of NLP task implemented by the system(s) evaluated (e.g.\ summarisation);
    \item Input/output language(s) used (e.g.\ English).
\end{enumerate}

\noindent During this first annotation, we manually filtered out papers only discussing human evaluation rather than including one (e.g., surveys of human evaluation), longitudinal studies, any that used highly specialised participants such as medical doctors,
and any that we roughly estimated to be too expensive for us to repeat (threshold \$2,000 in evaluator payments). This left 116 papers.
For these papers, 
Table~\ref{tab:first_anno_stats} in the appendix shows the counts\footnote{Because some papers include multiple properties, for example, multiple languages in machine translation systems, some rows will not sum to 116.} of the most common values for each property annotated.
English was  dominant as system language, 
used in over 90\% of papers.  The second most common language was Chinese, which was  used in just under 10\% of experiments. 
Language generation tasks were most common, with summarisation the most frequent task, followed by dialogue and MT.

About a third of papers did not specify  type of participant.  Among  papers that did specify this, 60\% used crowd-sourcing, with the vast majority of these being run on  Mechanical Turk.  It was generally difficult to find information about participants, with about half of papers not reporting the total number of participants. 
Very few papers included a clear description of the relationship between systems, data sets, items, and participants; number of judgments is therefore an estimate.

It became clear during  high-level annotation that fewer than 5\% of the 116  papers remaining after filtering were repeatable from publicly available information alone. Fundamental details like number and type of evaluators, instructions and training, and data evaluated are often omitted. Our next step was therefore to contact authors in the hope of obtaining the missing information. Lack of information about human evaluations has been commented on a number of times recently \cite{van-der-lee-etal-2019-best,howcroft-etal-2020-twenty,belz-etal-2020-disentangling}.

\subsection{More detailed annotation of experiments}\label{ssec:props}

In the next stage we carried out detailed annotation of evaluation properties preparatory to selecting a subset of such properties to control in our multi-lab multi-test study. We emailed the corresponding author (defaulting to first author) for each of the 116 papers to ask if they would support reproduction studies and, if they could provide more detailed information about their experiments.

The requested information included the user interface from the evaluation and the set of outputs shown to the evaluators (complete list see Appendix~\ref{sec:appendix:initial_author_information}). We received replies for just 39\%
of papers, even after sending reminders. Many of those who did reply were unable to provide the information needed. In the end, only 20 authors (20 papers containing 28 experiments) gave us enough information to progress the paper to the detailed annotation stage.\footnote{One further author did provide sufficient information, but upon further analysis of the paper and the resources they sent, we decided that the evaluation experiment reported in it was too different from the other 20 papers; the systems detected change in language use over time.}
The most common reason for authors responding but being unable to provide information was that they had moved on from their (usually graduate student) position and files had not been kept.  In some cases, authors from commercial research groups who were unable to provide information for business reasons. 
There were also eight papers where the authors responded initially, but the correspondence stalled.

Using the  author-provided information together with  paper, supplementary material and online resources, we annotated the 20 papers that progressed to this stage for the detailed properties of evaluations shown in   Section~\ref{sec:appendix:appendix_factor_annotation}, annotated at the level of individual experiments (28), 
because at this more fine-grained annotation level, properties can differ between different experiments in the same paper.

\begin{table}[t]
\setlength\tabcolsep{5.7pt} % default value: 6pt   
    \centering
    \begin{small}
    \begin{tabular}{|l|c|c|c|c}
        \hline
        \multirow{2}{*}{Training or expertise} & neither & only one & \multicolumn{2}{c|}{both} \\
        \cline{2-5}
         & 11 & 13 & \multicolumn{2}{c|}{4} \\
        \hline
        \hline
        \multirow{2}{*}{Number of participants} & \multicolumn{2}{c|}{small} & \multicolumn{2}{c|}{not small} \\
        \cline{2-5}
         & \multicolumn{2}{c|}{14} & \multicolumn{2}{c|}{14} \\
        \hline
        \hline
        \multirow{2}{*}{Complexity} & low & medium & \multicolumn{2}{c|}{high} \\
        \cline{2-5}
         & 9 & 11 & \multicolumn{2}{c|}{8} \\
        \hline
    \end{tabular}
    \caption{Frequency of control-factor annotations.}
    \label{tab:control-factors}
    \end{small}
\end{table}

One of the first three authors of the present paper annotated the 28 experiments with the detailed properties; the other two each checked half of the annotations.  Any differences were discussed and resolved.  To complete these annotations, we had to ask authors additional questions (usually in multiple rounds of questions and responses) for all experiments except two.  
In the end, for 8 of the 28 experiments we did not succeed in obtaining all the information needed for  the above properties.

Note that the last two properties in Section~\ref{sec:appendix:appendix_factor_annotation} (evaluation task complexity, interface complexity) have a different status from the others, in that they are secondary properties, subjectively assessed during annotation, rather than deriving from author-provided information. We found we tended to either agree on what their value should be, and when there was disagreement, values were adjacent. We used discussion rather than attempting to formalise rules to resolve disagreement, as it would seem an impossible task to exhaustively capture the latter.

Table~\ref{tab:control-factors}, and Table~\ref{tab:second-anno-stats} in the Appendix, show the frequency of the most common property values across the 28 experiments (here including unclear values).
We found that most of the annotated properties have one or two values that are the most frequent by large margins. For example, assessments were \textit{intrinsic} in 26 out of 28 experiments, \textit{subjective} in 26 out of 28, and \textit{absolute} in 20 out of 28. Only two experiments were \textit{extrinsic} and \textit{objective} evaluations, the other 26 were \textit{intrinsic} and \textit{subjective}.  There was large variation in the number of participants, with a low of 2 and a high of 233. None of the experiments provided explicit training sessions for participants, and only one included a practice session.  About three quarters of experiments provided instructions and/or criterion definitions.\footnote{We cannot be precise because this information was in some cases not provided even after we interacted with authors.} Around half of the experiments used subjects with specialist expertise, which was usually linguistics or NLP.

\subsection{Choosing properties to control for}\label{ssec:selecting-props}

The issues discussed in previous sections posed serious problems for selecting papers for a controlled study: we had only 20 fully annotated experiments; and we were left with very skewed distributions for many of the properties we had annotated, with many property combinations not occurring at all, or only occurring in one or two cases. Given the above issues it was clear that we were only going to be able to select a small set of properties to control for. 
We therefore whittled down the set of properties we had annotated to three that were both feasible and had a reasonable likelihood, based on existing work, of affecting reproducibility. For these, we created between two and three bins from the original value ranges, as follows:

\begin{enumerate}[itemsep=0pt]
    \item \textbf{\textit{Number of evaluators (small, not small)}}: Experiments with 1--5 evaluators were assigned the \textit{small} value, those with more than 5 evaluators the \textit{not small} value.
    
    \item \textbf{\textit{Cognitive complexity of assessment performed by evaluators (low, medium, high)}}: Experiments were assigned to one of the three possible values on the basis of the task complexity and interface complexity properties listed in Section~\ref{sec:appendix:appendix_factor_annotation}.
    
    \item \textbf{\textit{Training and/or expertise of evaluators (both, one, neither)}}: Experiments that had both trained, and required specific expertise from, evaluators were assigned \textit{both}; those that either trained evaluators or required expertise (but not both) were assigned \textit{one}; the remainder were assigned \textit{neither}.
\end{enumerate}

\begin{table*}[]
    \centering\small
    \begin{tabular}
    {|l|C{1.2cm}C{1.2cm}|C{1cm}C{1cm}C{1cm}|C{1cm}C{1cm}C{1cm}|}
        \hline
         & \multicolumn{2}{c|}{Num. Evaluators} & \multicolumn{3}{c|}{Cognitive Complexity} & \multicolumn{3}{c|}{Training and/or Experise} \\
        \cline{2-9}
        \multicolumn{1}{|p{1.5cm}|}{Task} & small & not small & low & medium & high & neither & either & both \\ \hline
        Dialogue & 1 & 0 & 0 & 1 & 0 & 0 & 1 & 0 \\
        Generation & 6 & 5 & 4 & 5 & 2 & 4 & 5 & 2 \\
        Summarisation & 3 & 1 & 2 & 1 & 1 & 1 & 3 & 0 \\
        Other & 2 & 2 & 1 & 0 & 3 & 2 & 0 & 2 \\ \hline
    \end{tabular}
    \caption{Counts of control property values per NLP task for the 20 experiments (from 15 papers) where all properties were clear.}
    \label{tab:per_task_factors}
\end{table*}

\noindent Even for this much reduced set of control factors, we did not have enough experiments to cover all 2$\times$3$\times$3 combinations of values, so we settled for a final set of 6 experiments, where there was an equal quantity of the pairwise combinations of the \textit{Number of evaluators} and \textit{Training/expertise} properties, as well as equal pairwise combinations of the \textit{Number of evaluators} and \textit{Complexity} properties.

\section{Setting up Reproductions}\label{sec:flaws}

Beginning the process of reproduction of the six experiments finally selected for reproduction (for common agreed approach to reproduction see Appendix~\ref{sec:common-appr}) necessarily involved delving into full implementational details for each of them. One particularly troubling finding has been the number of experimental flaws, errors and bugs we unearthed in the process. The more we dug into the properties of evaluation experiments that we needed in order to repeat an evaluation experiment, the more we uncovered flaws which made us question whether it made sense to repeat the experiment at all, in some cases because any conclusions drawn on the basis of the flawed experiments would be unsafe. Six specific issues are listed in Section~\ref{appsec:flaws}.\footnote{Note that we report these in anonymised form, because of the reputational risks involved. See also the Responsible Research Checklist included in the appendix.} Note that only one of our six selected experiments had none of these issues. We are still discovering more.

The structure we designed for our original study is shown in the Appendix Section~\ref{ssec:orig-study}, Figure~\ref{fig:orig-design}.

\section{Discussion}\label{sec:discussion}

The reasons why we decided to abandon our original study design were as follows. One, we struggled to find enough papers that did not have (i) prohibitive barriers to reproduction, and/or (ii) unavailable information that would be needed for repeating experiments, and/or (iii) experimental flaws and errors. Two, no matter how much effort we put into obtaining full experimental details from authors, there still remained questions, albeit increasingly fine-grained, that we did not have the answer to, such as if the presentation order of evaluated items was randomised, or what instructions/training participants were given. In some cases, information about additional things that had been done, but could not be guessed from previously provided information, transpired coincidentally, necessitating further changes to experimental design.

A potential solution to not having enough papers at the end is selecting more papers at the start (more years, more events). However, given the inordinate amount of work we put into obtaining enough information from authors,  simply tripling or quadrupling our initial pool of papers was not a viable solution. Similarly, there was little we were able to do about the reproduction barriers of excessive cost and highly specialised evaluators.

On the other hand, accepting to work from less than complete experimental information would have been problematic because information for different papers is incomplete in different ways, and we would not have been comparing like with like. 

Correcting flaws and errors would similarly have introduced differences between original and reproduction studies, moreover different ones in different cases. In this case we would strictly speaking no longer have been conducting reproductions.

We considered designing new evaluations from scratch with the properties we wanted for our MLMT study. However, it would have been very difficult to ensure that newly created studies were somehow representative of the kind of studies that are actually being conducted in NLP.

We have now opted for a solution incorporating elements from most of the above, where we select a somewhat larger set of existing studies in a process similar to before, reduce the number of different values of factors we control for, and then \textit{standardise and where necessary correct studies before reproduction}. Reproducibility is then measured between two new studies, rather than between them and the original study. 

\section{Conclusion}\label{sec:conclusion}

The track record of NLP as a field in recording information about human evaluation experiments is currently dire \cite{howcroft-etal-2020-twenty}. We saw in the paper-level
annotations (Appendix Table~\ref{tab:first_anno_stats}) that in 37 out of 116 papers the type of participant was unclear, in 59 the number of participants was unclear, and in 15 the number of judgements was unclear. Even after prolonged exchanges with authors during the experiment-level detailed annotation stage, very fundamental details were in some cases not obtainable: number of participants, details of training, instruction and practice items,  whether participants were required to be native speakers, and even the set of outputs evaluated.

Our overall conclusion is that, on the basis of the  unobtainability of information about experiments, barriers to reproduction and/or experimental flaws in our sample of 177 papers, only a small fraction of previous human evaluations in NLP can be repeated
under the same conditions, hence that their reproducibility cannot be tested by repeating them. The way forward would appear to be to accept the overhead of detailed recording of experimental details, e.g.\ with HEDS \cite{shimorina-belz-2022-human}, in combination with substantially increased standardisation in all aspects of experimental design.

\section*{Acknowledgements}

The ReproHum project is funded by EPSRC grant \href{https://gow.epsrc.ukri.org/NGBOViewGrant.aspx?GrantRef=EP/V05645X/1}{EP/V05645X/1}.  We would like to thank all authors who took the time to respond to our requests for information.  We would also like to thank Jackie Cheung.

\section*{Limitations}

The small subset of our findings that are based on information obtained from authors are necessarily limited in that they do not reflect information that might have been obtained from authors who did not respond.

Moreover, we selected our initial set of papers via search with key phrases ``human evaluation'' and ``participants.'' While this phrase is very commonly used to refer to non-automatic forms of evaluation, there is a chance that we may have missed papers because they used a different term. 

The small subset of conclusions based on our 
sample of  experiments are limited by their sample size in terms of how representative they are of current human evaluations in NLP more generally.

\section*{Ethics Statement}

As a paper that meta-reviews other academic publications, the present paper can be considered low-risk. Over and above collating information from publications, we annotated papers, analysed results and obtained descriptive statistics from annotations. In Section~5, we summarise the flaws, bugs and errors we found in experiments we were preparing for reproduction studies. We decided not to cite the papers where we found these, because the important information was that such issues occur, not which researchers were responsible for them.  

See also the responsible NLP research checklist completed for this paper (Appendix~\ref{ssec:responsible}).

\bibliography{mlmt-study}

\begin{thebibliography}{8}
\expandafter\ifx\csname natexlab\endcsname\relax\def\natexlab#1{#1}\fi

\bibitem[{Belz et~al.(2020)Belz, Mille, and
  Howcroft}]{belz-etal-2020-disentangling}
Anya Belz, Simon Mille, and David~M. Howcroft. 2020.
\newblock \href {https://aclanthology.org/2020.inlg-1.24} {Disentangling the
  properties of human evaluation methods: A classification system to support
  comparability, meta-evaluation and reproducibility testing}.
\newblock In \emph{Proceedings of the 13th International Conference on Natural
  Language Generation}, pages 183--194, Dublin, Ireland. Association for
  Computational Linguistics.

\bibitem[{Belz et~al.(2022)Belz, Popovic, and
  Mille}]{belz-etal-2022-quantified}
Anya Belz, Maja Popovic, and Simon Mille. 2022.
\newblock \href {https://doi.org/10.18653/v1/2022.acl-long.2} {Quantified
  reproducibility assessment of {NLP} results}.
\newblock In \emph{Proceedings of the 60th Annual Meeting of the Association
  for Computational Linguistics (Volume 1: Long Papers)}, pages 16--28, Dublin,
  Ireland. Association for Computational Linguistics.

\bibitem[{Howcroft et~al.(2020)Howcroft, Belz, Clinciu, Gkatzia, Hasan,
  Mahamood, Mille, van Miltenburg, Santhanam, and
  Rieser}]{howcroft-etal-2020-twenty}
David~M. Howcroft, Anya Belz, Miruna-Adriana Clinciu, Dimitra Gkatzia, Sadid~A.
  Hasan, Saad Mahamood, Simon Mille, Emiel van Miltenburg, Sashank Santhanam,
  and Verena Rieser. 2020.
\newblock \href {https://aclanthology.org/2020.inlg-1.23} {Twenty years of
  confusion in human evaluation: {NLG} needs evaluation sheets and standardised
  definitions}.
\newblock In \emph{Proceedings of the 13th International Conference on Natural
  Language Generation}, pages 169--182, Dublin, Ireland. Association for
  Computational Linguistics.

\bibitem[{Huidrom et~al.(2022)Huidrom, Du{\v{s}}ek, Kasner, Castro~Ferreira,
  and Belz}]{huidrom-etal-2022-two}
Rudali Huidrom, Ond{\v{r}}ej Du{\v{s}}ek, Zden{\v{e}}k Kasner, Thiago
  Castro~Ferreira, and Anya Belz. 2022.
\newblock \href {https://aclanthology.org/2022.inlg-genchal.9} {Two
  reproductions of a human-assessed comparative evaluation of a semantic error
  detection system}.
\newblock In \emph{Proceedings of the 15th International Conference on Natural
  Language Generation: Generation Challenges}, pages 52--61, Waterville, Maine,
  USA and virtual meeting. Association for Computational Linguistics.

\bibitem[{Pineau(2020)}]{pineau2020checklist}
Joelle Pineau. 2020.
\newblock \href
  {https://www.cs.mcgill.ca/~jpineau/ReproducibilityChecklist.pdf} {The machine
  learning reproducibility checklist v2.0}.

\bibitem[{Shimorina and Belz(2022)}]{shimorina-belz-2022-human}
Anastasia Shimorina and Anya Belz. 2022.
\newblock \href {https://doi.org/10.18653/v1/2022.humeval-1.6} {The human
  evaluation datasheet: A template for recording details of human evaluation
  experiments in {NLP}}.
\newblock In \emph{Proceedings of the 2nd Workshop on Human Evaluation of NLP
  Systems (HumEval)}, pages 54--75, Dublin, Ireland. Association for
  Computational Linguistics.

\bibitem[{van~der Lee et~al.(2019)van~der Lee, Gatt, van Miltenburg, Wubben,
  and Krahmer}]{van-der-lee-etal-2019-best}
Chris van~der Lee, Albert Gatt, Emiel van Miltenburg, Sander Wubben, and Emiel
  Krahmer. 2019.
\newblock \href {https://doi.org/10.18653/v1/W19-8643} {Best practices for the
  human evaluation of automatically generated text}.
\newblock In \emph{Proceedings of the 12th International Conference on Natural
  Language Generation}, pages 355--368, Tokyo, Japan. Association for
  Computational Linguistics.

\bibitem[{Whitaker(2017)}]{whitaker2017}
Kirstie Whitaker. 2017.
\newblock The {MT} {R}eproducibility {C}hecklist.
\newblock
  \url{{https://www.cs.mcgill.ca/~jpineau/ReproducibilityChecklist.pdf}}.

\end{thebibliography}
\bibliographystyle{acl_natbib}

\newpage

\appendix

\section{Appendix}
\label{sec:appendix}

\subsection{Original study design}\label{ssec:orig-study}

Figure~\ref{fig:orig-design} shows the original design of the multi-lab multi-test study.

\begin{figure*}[h!t]
\small 
\begin{flushleft}{Structural design for a multi-lab, multi-test controlled study of experimental factors affecting reproducibility:}
\end{flushleft}
\vspace{.3cm}
    \begin{centering}
\begin{enumerate}[itemsep=0.1pt,topsep=.3pt,labelwidth=0cm,itemindent=-.5cm]
    \item[]\textbf{\textit{Round 1:}} Testing precision under repeatability conditions of measurement.
    \begin{itemize}[leftmargin=0.3cm,itemsep=0pt,topsep=.2pt]
        \item Reproductions per experiment: 2 by two different labs;
        \item Conditions (experimental factors) to vary: evaluator cohort; 
        \item If reproduction close enough, go to Round~2, else repeat Round~1 with improvements to experimental design, in terms of increased number of evaluators, and decreased cognitive complexity of evaluation task;
        \item For Round~1 repeats, if reproducibility is increased between reproduction studies (compared to each other, not the original study), proceed to Round~2, else stop.
    \end{itemize}
    \item[]\textbf{\textit{Round 2:}} Testing reproducibility under varied conditions.
    \begin{itemize}[leftmargin=0.3cm,itemsep=0pt,topsep=.2pt]
        \item Reproductions per experiment: 2 by two different labs;
        \item Conditions (experimental factors) to vary: evaluator cohort, and either number of evaluators \textit{or} task complexity; 
        \item If reproduction close enough, go to Round~3, else repeat Round~2 with improvements to experimental design, in terms of increased number of evaluators, and decreased cognitive complexity of evaluation task. 
        \item For Round~2 repeats, if reproducibility is increased between reproduction studies (compared to each other, not the original study), proceed to Round~3, else stop.
    \end{itemize}
    \item[]\textbf{\textit{Round 3:}} Testing reproducibility under increasingly varied conditions.
    \begin{itemize}[leftmargin=0.3cm,itemsep=0pt,topsep=.2pt]
        \item Reproductions per experiment: 2 by two different labs;
        \item Conditions (experimental factors) to vary: evaluator cohort, number of evaluators \textit{and}  complexity.
    \end{itemize}
\end{enumerate} 
    \end{centering}
    \caption{Original design for the multi-lab, multi-test controlled study with a set of original human evaluation experiments with balanced experimental factors.}
    \label{fig:orig-design}
\end{figure*}

\subsection{Initial information requested from authors}
\label{sec:appendix:initial_author_information}

Our initial email to authors asked if they would be able to provide the following information:

\begin{enumerate}[itemsep=0pt]
    \item The system outputs that were shown to participants.
    \item The interface, form, or document that participants completed; the exact document or form that was used would be ideal.
    \item Details on the number and type of participants (students, researchers, Mechanical Turk, etc.) that took part in the study.
    \item The total cost of the original study.
\end{enumerate}

\subsection{Counts for high-level annotations}
\label{sec:high-level-annos}

Table~\ref{tab:first_anno_stats} shows counts for the first round of annotating paper-level properties.

\begin{table*}[]
    \centering
    \begin{small}
    \begin{tabular}{|p{3.8cm}|C{2.95cm}|C{2.95cm}|C{2.95cm}|C{1.1cm}|}
        \hline
		\multirow{2}{*}{System language(s)} & \cellcolor{propertyshade} \textit{English} & \cellcolor{propertyshade} \textit{Chinese} & \cellcolor{propertyshade} \textit{German} & \cellcolor{propertyshade} \textit{other} \\
		\cline{2-5}
		& 109 & 11 & 9 & 5\\
		\hline
		\hline
		\multirow{2}{*}{NLP Task} & \cellcolor{propertyshade} \textit{summarisation} & \cellcolor{propertyshade} \textit{dialogue systems} & \cellcolor{propertyshade} \textit{machine translation} & \cellcolor{propertyshade} \textit{other} \\
		\cline{2-5}
		& 33 & 22 & 9 & 55\\
		\hline
		\hline
		\multirow{2}{*}{Number of systems} & \cellcolor{propertyshade} \textit{1-5} & \cellcolor{propertyshade} \textit{6-7} & \cellcolor{propertyshade} \textit{> 7} & \cellcolor{propertyshade} \textit{unclear} \\
		\cline{2-5}
		& 89 & 14 & 13 & 0\\
		\hline
		\hline
		\multirow{2}{*}{Number of datasets} & \cellcolor{propertyshade} \textit{1} & \cellcolor{propertyshade} \textit{2} & \cellcolor{propertyshade} \textit{> 3} & \cellcolor{propertyshade} \textit{unclear} \\
		\cline{2-5}
		& 83 & 25 & 8 & 0\\
		\hline
		\hline
		\multirow{2}{*}{Type of participant} & \cellcolor{propertyshade} \textit{crowd (e.g., MTurk)} & \cellcolor{propertyshade} \textit{author/colleague/student} & \cellcolor{propertyshade} \textit{other} & \cellcolor{propertyshade} \textit{unclear} \\
		\cline{2-5}
		& 47 & 21 & 14 & 37\\
		\hline
		\hline
		\multirow{2}{*}{Number of unique participants} & \cellcolor{propertyshade} \textit{< 5} & \cellcolor{propertyshade} \textit{5-20} & \cellcolor{propertyshade} \textit{> 20} & \cellcolor{propertyshade} \textit{unclear} \\
		\cline{2-5}
		& 27 & 19 & 11 & 59\\
		\hline
		\hline
		\multirow{2}{*}{Number of judgments} & \cellcolor{propertyshade} \textit{< 100} & \cellcolor{propertyshade} \textit{100-1000} & \cellcolor{propertyshade} \textit{> 1000} & \cellcolor{propertyshade} \textit{unclear} \\
		\cline{2-5}
		& 1 & 34 & 66 & 15\\
		\hline

    \end{tabular}
    \caption{Frequency of the high-level experimental properties in the 116 papers, at the paper level.  Some papers have multiple categorical properties therefore some rows will not sum to 116.}
    \label{tab:first_anno_stats}
    \end{small}
\end{table*}

\subsection{Details of experiment-level annotation}
\label{sec:appendix:appendix_factor_annotation}

\begin{table*}[t]
    \centering
    \begin{small}
    \begin{tabular}{|p{3.8cm}|cccccccccccc|}
        \hline
        \multirow{2}{*}{Quality criteria names} & \multicolumn{3}{C{2.5cm}|}{\cellcolor{propertyshade} \textit{fluency}} & \multicolumn{3}{C{2.5cm}|}{\cellcolor{propertyshade} \textit{coherence}} & \multicolumn{3}{C{2.5cm}|}{\cellcolor{propertyshade} \textit{informativeness}} & \multicolumn{3}{C{2.5cm}|}{\cellcolor{propertyshade} \textit{other}} \\ \cline{2-13} 
        & \multicolumn{3}{C{2.5cm}|}{10} & \multicolumn{3}{C{2.5cm}|}{5} & \multicolumn{3}{C{2.5cm}|}{3} & \multicolumn{3}{C{2.5cm}|}{54} \\
        \hline
        \hline
        \multirow{2}{*}{System language(s)} & \multicolumn{3}{C{2.5cm}|}{\cellcolor{propertyshade} \textit{English}} & \multicolumn{3}{C{2.5cm}|}{\cellcolor{propertyshade} \textit{Chinese}} & \multicolumn{3}{C{2.5cm}|}{\cellcolor{propertyshade} \textit{German}} & \multicolumn{3}{C{2.5cm}|}{\cellcolor{propertyshade} \textit{other}} \\ \cline{2-13} 
        & \multicolumn{3}{C{2.5cm}|}{26} & \multicolumn{3}{C{2.5cm}|}{3} & \multicolumn{3}{C{2.5cm}|}{2} & \multicolumn{3}{C{2.5cm}|}{0} \\
        \hline
       \hline
        \multirow{2}{*}{NLP Task} & \multicolumn{3}{C{2.5cm}|}{\cellcolor{propertyshade} \textit{summarisation}} & \multicolumn{3}{C{2.5cm}|}{\cellcolor{propertyshade} \textit{question answering}} & \multicolumn{3}{C{2.5cm}|}{\cellcolor{propertyshade} \textit{explanation}} & \multicolumn{3}{C{2.5cm}|}{\cellcolor{propertyshade} \textit{other}} \\ \cline{2-13} 
        & \multicolumn{3}{C{2.5cm}|}{6} & \multicolumn{3}{C{2.5cm}|}{3} & \multicolumn{3}{C{2.5cm}|}{3} & \multicolumn{3}{C{2.5cm}|}{16} \\
        \hline
        \hline
        \multirow{2}{*}{Type of participant} & \multicolumn{3}{C{2.5cm}|}{\cellcolor{propertyshade} \textit{crowd}} & \multicolumn{3}{C{2.5cm}|}{\cellcolor{propertyshade} \textit{student}} & \multicolumn{3}{C{2.5cm}|}{\cellcolor{propertyshade} \textit{colleague}} & \multicolumn{3}{C{2.5cm}|}{\cellcolor{propertyshade} \textit{other}} \\ \cline{2-13} 
        & \multicolumn{3}{C{2.5cm}|}{13} & \multicolumn{3}{C{2.5cm}|}{8} & \multicolumn{3}{C{2.5cm}|}{7} & \multicolumn{3}{C{2.5cm}|}{4} \\
        \hline
        \hline
        & \multicolumn{6}{c|}{\cellcolor{propertyshade} \textit{intrinsic}} & \multicolumn{6}{c|}{\cellcolor{propertyshade} \textit{extrinsic}} \\ \cline{2-13} 
        \multirow{-2}{*}{Intrinsic or extrinsic} & \multicolumn{6}{c|}{26} & \multicolumn{6}{c|}{2} \\
        \hline
        \hline
        & \multicolumn{6}{c|}{\cellcolor{propertyshade} \textit{absolute}} & \multicolumn{6}{c|}{\cellcolor{propertyshade} \textit{relative}} \\ \cline{2-13} 
        \multirow{-2}{*}{Absolute or relative} & \multicolumn{6}{c|}{20} & \multicolumn{6}{c|}{8} \\
        \hline
        \hline
        & \multicolumn{6}{c|}{\cellcolor{propertyshade} \textit{objective}} & \multicolumn{6}{c|}{\cellcolor{propertyshade} \textit{subjective}} \\ \cline{2-13} 
        \multirow{-2}{*}{Objective or subjective} & \multicolumn{6}{c|}{2} & \multicolumn{6}{c|}{26} \\
        \hline
        \hline
        \multirow{2}{*}{Num. of unique participants} & \multicolumn{3}{C{2.5cm}|}{\cellcolor{propertyshade} \textit{$<$ 5}} & \multicolumn{3}{C{2.5cm}|}{\cellcolor{propertyshade} \textit{5--20}} & \multicolumn{3}{C{2.5cm}|}{\cellcolor{propertyshade} \textit{$>$ 20}} & \multicolumn{3}{C{2.5cm}|}{\cellcolor{propertyshade} \textit{unclear}} \\ \cline{2-13} 
        & \multicolumn{3}{C{2.5cm}|}{11} & \multicolumn{3}{C{2.5cm}|}{4} & \multicolumn{3}{C{2.5cm}|}{8} & \multicolumn{3}{C{2.5cm}|}{5} \\
        \hline
        \hline
        \multirow{2}{*}{Num. of items evaluated} & \multicolumn{3}{C{2.5cm}|}{\cellcolor{propertyshade} \textit{$<$ 200}} & \multicolumn{3}{C{2.5cm}|}{\cellcolor{propertyshade} \textit{200--1000}} & \multicolumn{3}{C{2.5cm}|}{\cellcolor{propertyshade} \textit{$>$ 1000}} & \multicolumn{3}{C{2.5cm}|}{\cellcolor{propertyshade} \textit{unclear}} \\ \cline{2-13} 
        & \multicolumn{3}{C{2.5cm}|}{9} & \multicolumn{3}{C{2.5cm}|}{10} & \multicolumn{3}{C{2.5cm}|}{7} & \multicolumn{3}{C{2.5cm}|}{2} \\
        \hline
        \hline
        \multirow{2}{*}{Num. of participants per item} & \multicolumn{3}{C{2.5cm}|}{\cellcolor{propertyshade} \textit{$<$ 4}} & \multicolumn{3}{C{2.5cm}|}{\cellcolor{propertyshade} \textit{4--9}} & \multicolumn{3}{C{2.5cm}|}{\cellcolor{propertyshade} \textit{$>$ 9}} & \multicolumn{3}{C{2.5cm}|}{\cellcolor{propertyshade} \textit{varies}} \\ \cline{2-13} 
        & \multicolumn{3}{C{2.5cm}|}{17} & \multicolumn{3}{C{2.5cm}|}{3} & \multicolumn{3}{C{2.5cm}|}{3} & \multicolumn{3}{C{2.5cm}|}{5} \\
        \hline
        \hline
        \multirow{2}{*}{Num. of items per participant} & \multicolumn{3}{C{2.5cm}|}{\cellcolor{propertyshade} \textit{$<$ 50}} & \multicolumn{3}{C{2.5cm}|}{\cellcolor{propertyshade} \textit{50--200}} & \multicolumn{3}{C{2.5cm}|}{\cellcolor{propertyshade} \textit{$>$ 200}} & \multicolumn{3}{C{2.5cm}|}{\cellcolor{propertyshade} \textit{varies/unclear}} \\ \cline{2-13} 
        & \multicolumn{3}{C{2.5cm}|}{5} & \multicolumn{3}{C{2.5cm}|}{5} & \multicolumn{3}{C{2.5cm}|}{7} & \multicolumn{3}{C{2.5cm}|}{11} \\
        \hline
        \hline
        & \multicolumn{6}{c|}{\cellcolor{propertyshade} \textit{no}} & \multicolumn{6}{c|}{\cellcolor{propertyshade} \textit{unclear}} \\ \cline{2-13} 
        \multirow{-2}{*}{Training given} & \multicolumn{6}{c|}{24} & \multicolumn{6}{c|}{4} \\
        \hline
        \hline
         & \multicolumn{4}{C{3.5cm}|}{\cellcolor{propertyshade} \textit{yes}} & \multicolumn{4}{C{3.5cm}|}{\cellcolor{propertyshade} \textit{no}} & \multicolumn{4}{C{3.5cm}|}{\cellcolor{propertyshade} \textit{unclear}} \\ \cline{2-13} 
        \multirow{-2}{*}{Instructions given} & \multicolumn{4}{c|}{8} & \multicolumn{4}{c|}{15} & \multicolumn{4}{c|}{5} \\
        \hline
        \hline
        \multirow{2}{*}{Criterion definitions given} & \multicolumn{3}{C{2.5cm}|}{\cellcolor{propertyshade} \textit{yes}} & \multicolumn{3}{C{2.5cm}|}{\cellcolor{propertyshade} \textit{no}} & \multicolumn{3}{C{2.5cm}|}{\cellcolor{propertyshade} \textit{n/a}} & \multicolumn{3}{C{2.5cm}|}{\cellcolor{propertyshade} \textit{unclear/mixed}} \\ \cline{2-13} 
        & \multicolumn{3}{C{2.5cm}|}{17} & \multicolumn{3}{C{2.5cm}|}{3} & \multicolumn{3}{C{2.5cm}|}{4} & \multicolumn{3}{C{2.5cm}|}{4} \\
        \hline
        \hline
         & \multicolumn{4}{C{3.5cm}|}{\cellcolor{propertyshade} \textit{yes}} & \multicolumn{4}{C{3.5cm}|}{\cellcolor{propertyshade} \textit{no}} & \multicolumn{4}{C{3.5cm}|}{\cellcolor{propertyshade} \textit{unclear}} \\ \cline{2-13} 
        \multirow{-2}{*}{Practice session held} & \multicolumn{4}{c|}{1} & \multicolumn{4}{c|}{23} & \multicolumn{4}{c|}{4} \\
        \hline
        \hline
        \multirow{2}{*}{Participant expertise type} & \multicolumn{3}{C{2.5cm}|}{\cellcolor{propertyshade} \textit{none}} & \multicolumn{3}{C{2.5cm}|}{\cellcolor{propertyshade} \textit{researcher}} & \multicolumn{3}{C{2.5cm}|}{\cellcolor{propertyshade} \textit{linguist}} & \multicolumn{3}{C{2.5cm}|}{\cellcolor{propertyshade} \textit{domain}} \\ \cline{2-13} 
        & \multicolumn{3}{C{2.5cm}|}{16} & \multicolumn{3}{C{2.5cm}|}{9} & \multicolumn{3}{C{2.5cm}|}{2} & \multicolumn{3}{C{2.5cm}|}{1} \\
        \hline
        \hline
        \multirow{2}{*}{Participants native speakers} & \multicolumn{3}{C{2.5cm}|}{\cellcolor{propertyshade} \textit{yes}} & \multicolumn{3}{C{2.5cm}|}{\cellcolor{propertyshade} \textit{no}} & \multicolumn{3}{C{2.5cm}|}{\cellcolor{propertyshade} \textit{of region}} & \multicolumn{3}{C{2.5cm}|}{\cellcolor{propertyshade} \textit{unknown}} \\ \cline{2-13} 
        & \multicolumn{3}{C{2.5cm}|}{2} & \multicolumn{3}{C{2.5cm}|}{12} & \multicolumn{3}{C{2.5cm}|}{10} & \multicolumn{3}{C{2.5cm}|}{4} \\
        \hline
    \end{tabular}
    \caption{Frequency of detailed experimental properties in set of 28 experiments.}
    \label{tab:second-anno-stats}
    \end{small}
\end{table*}

All of the property names and values from our detailed annotations are listed below, along with descriptions of what was recorded for each property:

\begin{enumerate}[itemsep=0pt]
    \item Specific data sets used;
    \item Specific evaluation criteria names used;  the criterion names as stated in the paper if possible, otherwise a criterion name that represents what is being assessed.
    \item System languages;  the language(s) used by the system as either input or output.
    \item System task;  the NLP task that the system is tackling.  Values from the 28 experiments were cross-lingual summarisation, data-to-text generation, definition generation with controllable complexity, dialogue summarisation, dialogue turn generation, explanation generation, fact-check justification generation, machine translation error prediction, prompted generation, question generation, question-answer generation, referring expression generation, simplification, summarisation, text to speech.
    \item Evaluator type;  the type of evaluator, values included colleagues, commercial in-house evaluators, crowd-sourced, mix of author and colleague, mix of colleague and students, professional, student.
    \item Evaluation modes \cite{belz-etal-2020-disentangling}:
        \begin{enumerate}[itemsep=0pt]
            \item Intrinsic vs.\ extrinsic;
            \item Absolute vs.\ relative;
            \item Objective vs.\ subjective.
        \end{enumerate}
    \item Number of participants;  the total number of unique participants that took part in the study,
    \item Number of items evaluated;  in the case of an absolute evaluation this is one system output.  In the case of a relative evaluation, it refers to the set of outputs, e.g., a pair, that is being compared.
    \item How many participants evaluated each item;  for some experiments, this varied.
    \item How many items were evaluated by each participant;  for some experiments, this varied.  In particular, for the 13 of 28 experiments that were crowd-sourced, 5 were known integers, 4 varied, and 4 could not be determined (we suspect these also varied).
    \item Were training and/or practice sessions provided for participants;  see the discussion below.
    \item Were participants given instructions?  Were they given definitions of evaluation criteria;  see the discussion below.
    \item Were participants required to have a specific expertise?  If so, what type, and was this self-reported or externally assessed?;  see the discussion below.
    \item Were participants required to be native speakers?  If so, was this self-reported or externally assessed?;  For the first part we used the options yes, no, crowd-source region filters, and in one case that the experiment was performed with students at a university where the language was native.    The latter two are inherently self-reported, although with some limited control by the researchers.  Only for one of the experiments with native speakers did the researchers indicate that they had confirmed this, all others were self-reports.
    \item How complex was the evaluation task (low, medium, high);  assessment by authors of this paper.
    \item How complex was the interface (low, medium, high);  assessment by authors of this paper.
\end{enumerate}

\noindent Classifying the type of participant, training, instruction, and expertise was very difficult.  Firstly, not all experiments necessarily require detailed instructions but setting a threshold beyond which instructions become non-perfunctory is difficult.  The same is true for training.  In the end, we decided to record whether there non-perfunctory training, instruction, practice, or criterion definition.

Expertise was also difficult to classify.  Some papers would have originally reported `expert annotators', but following our queries stated participants were graduate students or colleagues.  Such participants were often called `NLP experts'.  In the end, we considered participants to be expert if the authors of the original study indicated that they were.

\subsection{Common Approach to Reproduction}\label{sec:common-appr}

In order to ensure comparability between studies, we agreed the following common-ground approach to carrying out reproduction studies:

\begin{enumerate}\itemsep=0pt
    \item Plan for repeating the original experiment identically, then apply to research ethics committee for approval.
    \item If participants were paid during the original experiment, determine pay in accordance with the common procedure for calculating fair pay (see appendix).
    
    \item Complete HEDS datasheet.
    \item Identify the following types of results reported in the original paper for the experiment:
    \begin{enumerate}
        \item Type I results: single numerical scores, e.g. mean quality rating, error count, etc.
        \item Type II results: sets of numerical scores, e.g. set of Type I results.
        \item Type III results: categorical labels attached to text spans of any length.
        \item Qualitative conclusions/findings stated explicitly in the original paper.
    \end{enumerate}
    
    \item Carry out the allocated experiment exactly as described in the HEDS sheet.
    \item Report quantified reproducibility assessments for 8a--c as follows:
        \begin{enumerate}
        \item Type I results:  Coefficient of variation (debiased for small samples).
        \item Type II results: Pearson’s r, Spearman’s $\rho$.
        \item Type III results: Multi-rater: Fleiss’s $\kappa$; Multi-rater, multi-label: Krippendorff’s $\alpha$.
        \item Conclusions/findings: 
        Side-by-side summary of conclusions/findings that are / are not confirmed in the repeat experiment.
    \end{enumerate}
\end{enumerate}

\subsection{Issues, flaws and errors found}\label{appsec:flaws}
\begin{enumerate}\itemsep=0pt
    \item Mistakes in the reported figures for the human evaluation in the published paper, with the result that systems were reported as being better or worse that they actually were.
    \item Reporting a total number of items in the paper which did not match the files that were sent. % 
    \item Failure to randomise the order of items to be evaluated (when the stated intention was to randomise) due to wrongly applied randomisation. 
    \item Reporting that evaluators did equal numbers of assessments but it's clear from the files that they did very different numbers.
    \item Ad-hoc attention checks (exact nature of which authors were unable to provide) applied to some but not all participants who if they failed the check were excluded from further contributing to the experiment, but whose already completed work was kept.
    \item{Biased methods of aggregating judgments (choosing a preferred participant rather than using some form of average).}
\end{enumerate}

\noindent On a more general note, ambiguities in the reporting can be an issue. Even when checked against the HEDS sheet, authors could feel like they have mentioned all experimental details that are asked for in HEDS, but often these are described at such a high level that there is still room for misinterpretation, which means that authors still need to confirm that their paper has been interpreted correctly. One solution for NLP authors could be to let a third party fill in the HEDS sheet and see where they get stuck, but this does add a further overhead.

\subsection{ARR Responsible Research Checklist}\label{ssec:responsible}

\begin{itemize}
    \item[A.] \textbf{For every submission:}
    \begin{itemize}
        \item[A1.] \textbf{Did you describe the limitations of your work? }Yes, e.g. we discuss the limitations from having a self-selecting subset of papers (where authors responded) available for analysis rather than a complete one.
    \item[A2.] \textbf{Did you discuss any potential risks of your work?}
The work analyses previously peer-reviewed and published human evaluation experiments, and while conventional risk considerations don’t apply, we do mention the potential harm to individual authors from non-anonymously reporting experimental flaws and/or low reproducibility in their work.
    \item[A3.] \textbf{Do the abstract and introduction summarise the paper’s main claims?}
Yes, abstract, introduction and conclusion  summarise main aims and conclusions from the work.
    \end{itemize}
    \item[B.] \textbf{Did you use or create scientific artefacts?}
No new data or computational resources were created.
\item[C.] \textbf{Did you run computational experiments?}
No experiments were run.
\item[D.] \textbf{Did you use human annotators (e.g., crowdworkers) or research with human participants?}
No human annotation or evaluations were carried out for this paper (other than by the authors).
\end{itemize}

\end{document}